\documentclass{article}

\usepackage{arxiv}

\usepackage{subcaption}
\usepackage{amsmath}
\usepackage{appendix}

\usepackage[utf8]{inputenc} 
\usepackage[T1]{fontenc}    
\usepackage{hyperref}       
\usepackage{url}            
\usepackage{booktabs}       
\usepackage{amsfonts}       
\usepackage{nicefrac}       
\usepackage{microtype}      
\usepackage{cleveref}       
\usepackage{graphicx}
\usepackage{natbib}
\usepackage{doi}
\usepackage{float}

\title{Reliable Evaluation of Attribution Maps in CNNs: A Perturbation-Based Approach}

\author{ \href{https://orcid.org/0000-0002-7523-5694}{\includegraphics[scale=0.06]{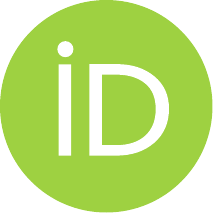}\hspace{1mm}Lars~Nieradzik}\\
	Image Processing Department\\
	Fraunhofer ITWM\\
	Fraunhofer Platz 1, 67663, Kaiserslautern\\
	\texttt{lars.nieradzik@itwm.fraunhofer.de}\\
	\And
	\href{https://orcid.org/0000-0002-9821-1636}{\includegraphics[scale=0.06]{orcid.pdf}\hspace{1mm}Henrike~Stephani} \\
	Image Processing Department\\
	Fraunhofer ITWM\\
	Fraunhofer Platz 1, 67663, Kaiserslautern\\
	\texttt{henrike.stephani@itwm.fraunhofer.de}\\
	\And
	\href{https://orcid.org/0000-0002-1327-1243}{\includegraphics[scale=0.06]{orcid.pdf}\hspace{1mm}Janis~Keuper} \\
	Institute of Machine Learning and Analysis (IMLA)\\
	Offenburg University\\
	Badstr. 24, 77652, Offenburg\\
	\texttt{jkeuper@ad.hs-offenburg.de}\\
}

\renewcommand{\shorttitle}{Reliable Evaluation of Attribution Maps in CNNs}

\hypersetup{
pdftitle={A template for the arxiv style},
pdfsubject={q-bio.NC, q-bio.QM},
pdfauthor={David S.~Hippocampus, Elias D.~Striatum},
pdfkeywords={First keyword, Second keyword, More},
}

\begin{document}
\maketitle

\begin{abstract}In this paper, we present an approach for evaluating attribution maps, which play a central role in interpreting the predictions of convolutional neural networks (CNNs). We show that the widely used insertion/deletion metrics are susceptible to distribution shifts that affect the reliability of the ranking. Our method proposes to replace pixel modifications with adversarial perturbations, which provides a more robust evaluation framework. By using smoothness and monotonicity measures, we illustrate the effectiveness of our approach in correcting distribution shifts. In addition, we conduct the most comprehensive quantitative and qualitative assessment of attribution maps to date. Introducing baseline attribution maps as sanity checks, we find that our metric is the only contender to pass all checks. Using Kendall's $\tau$ rank correlation coefficient, we show the increased consistency of our metric across 15 dataset-architecture combinations. Of the 16 attribution maps tested, our results clearly show SmoothGrad to be the best map currently available. This research makes an important contribution to the development of attribution maps by providing a reliable and consistent evaluation framework. To ensure reproducibility, we will provide the code along with our results.
\end{abstract}



\section{Introduction}
The explainability of learned models is a crucial, yet mostly unachieved property towards a general utilization and acceptance of machine learning techniques \cite{gilpin2018explaining}. In order to succeed, many practical applications require solutions that provide not only high test accuracy, but also robustness, uncertainty estimates for predictions, and ways to make decision processes understandable for humans. Although recent deep learning methods have been very successful in achieving great advances in terms of model accuracy on many different tasks \cite{zarandy2015overview}, the other properties are mostly still the subject of increasing research efforts.

\begin{figure}[t]
  \centering
  \includegraphics[scale=0.18]{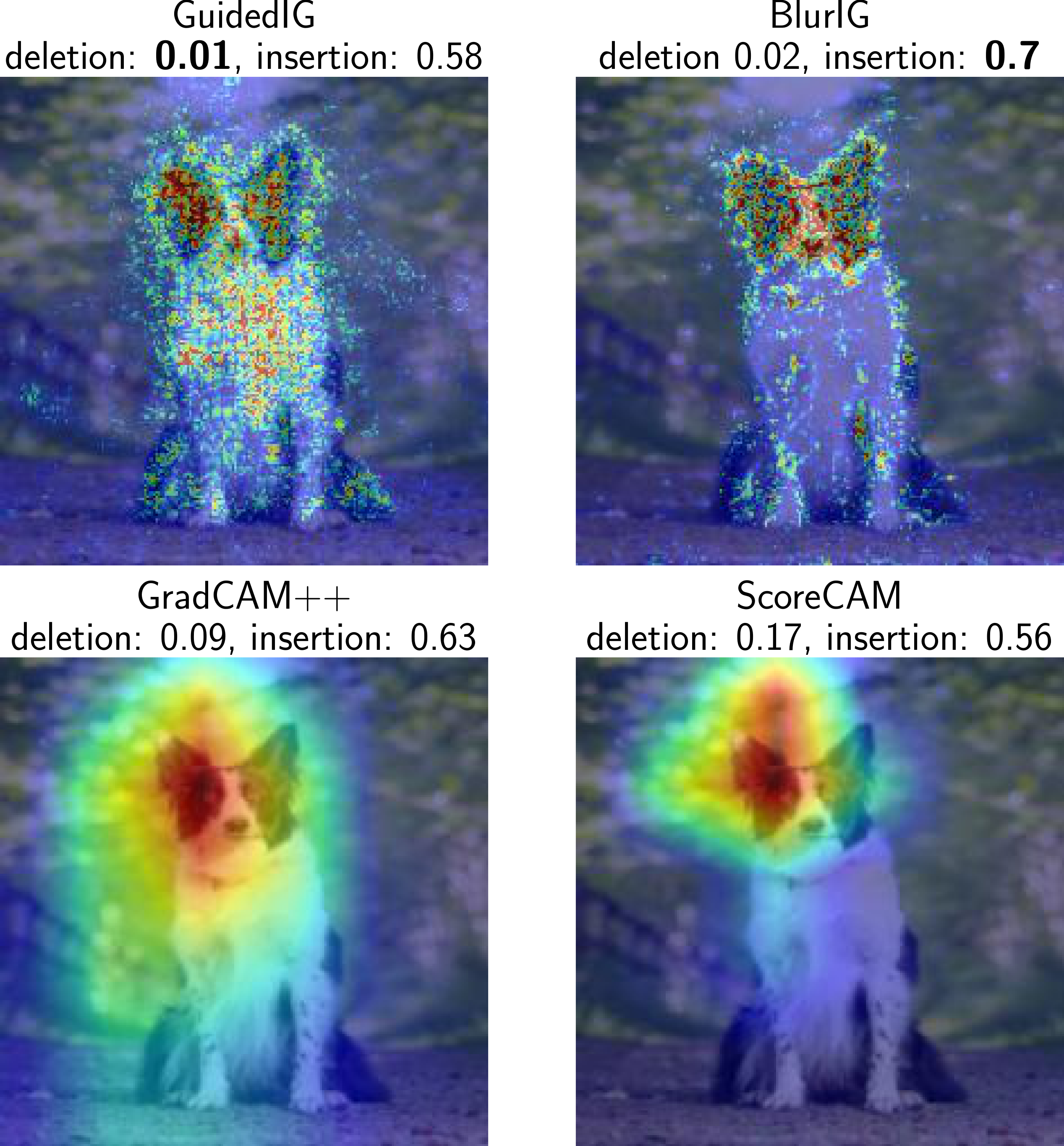}
  
   \caption{Visualization of four different attribution maps (AM) and two evaluation methods on the same input image and the same model (EfficientNet-B0 \cite{DBLP:journals/corr/abs-1905-11946}). Visually, it is impossible to objectively determine which result gives the ``best'' estimate of the image regions with the highest impact on the model decision. Current evaluation methods of the used AMs also give conflicting results: the {\it Deletion} method (lower is better) \cite{DBLP:journals/corr/FongV17, DBLP:journals/corr/abs-1806-07421} ranks {\it Guided Integrated Gradients (IG)} \cite{DBLP:journals/corr/abs-2106-09788} first,  while the {\it Insertion} method (higher is better)  points towards {\it Blur IG} \cite{DBLP:journals/corr/abs-2004-03383}. Refer also to \cref{fig:corr} for a comparison between more images.}
   \label{fig:zero}
\end{figure}
A current standard tool for the visualization of decision processes in neural networks are so-called {\it Attribution Methods} \cite{zhang2021survey}. In the case of {\it Computer Vision} tasks with image input data, attribution methods are used to compute {\it Attribution Maps} (AM) (also known as {\it Saliency Maps} \cite{zhang2021survey}) of individual input images on a given trained model. AMs provide a human-interpretable visualization of the image regions' weighted impact on the model, allowing intuitive explanations of the complex internal mappings of model predictions. However, the qualitative nature of AMs makes it very hard to validate their actual correctness or to define a performance measure for AM methods \cite{nieradzik2024challenging}. Fig \ref{fig:zero} gives a representative example of the large variations in the visual results of four different prominent AM algorithms, intuitively showing the need for quantitative AM evaluation methods in terms of the following, practical questions: \textbf{i) How can we evaluate the correctness of AM outputs? ii) How can we compare the performance of the many recently published AM algorithms? iii) Which attribution method should we use for the investigation of a given research question or development goal? }\\
  
To address these questions, we propose a novel quantitative evaluation method for AMs in {\it Convolutional Neural Networks} (CNN) \cite{lecun1995convolutional}, utilizing adversarial perturbations to verify the given importance of image regions to CNN predictions by attribution methods. The key contributions of this paper are:

\begin{itemize}
    \item We present a novel approach for attribution map evaluation that addresses the distributional shift problem encountered in widely used insertion/deletion metrics. Our method uses adversarial perturbations instead of pixel modifications, resulting in a more robust evaluation framework.
    
    \item We perform the most comprehensive quantitative and qualitative evaluation of attribution maps to date, covering 16 attribution methods across 15 dataset-architecture combinations.
    
    \item We demonstrate the increased consistency and reliability of our metric using Kendall's rank correlation coefficient $\tau$, baseline attribution maps, smoothness, monotonicity and a qualitative analysis.
\end{itemize}

\section{Related Work}

\subsection{Attribution methods}

Since the literature on attribution methods has become extensive in recent years, and many approaches are based on earlier methods, we retreat to the discussion and evaluation of some of the most widely used AM techniques in the context of CNNs.

\textbf{Full back-propagation methods}, like \emph{Gradients} \cite{DBLP:journals/corr/SimonyanVZ13}, have been the basis for the first AM approaches for classification models: they compute a gradient on a learned neural network $f$ with respect to a given input $X$ for a fixed target class $c$:  $\frac{\partial f^{(c)}}{\partial X}$. The derivation of this approach becomes quite intuitive, when we retreat to a simple linear decision function $f(x) = w^Tx$. Then, the weights $w$ are a direct indicator of the input importance, as  we have $\frac{\partial f}{\partial x} = w$.
The later introduced \emph{DeConvNet} \cite{DBLP:journals/corr/ZeilerF13} and \emph{Guided backpropagation} \cite{DBLP:journals/corr/SpringenbergDBR14} modify the gradients of the widely used {\it ReLU} activation to enable the backward flow of negative gradients. Another extension of \emph{Gradients} has been introduced by {\it SmoothGrad} \cite{DBLP:journals/corr/SmilkovTKVW17}, which computes the gradient multiple times by adding Gaussian noise to the input and averaging the results.

\textbf{Path backpropagation methods} parameterize a path from a fixed baseline image to the given input image and compute derivatives analogous to {\it Gradients} along this path. For example, \emph{Integrated Gradients} \cite{DBLP:journals/corr/SundararajanTY17} parameterizes the path as straight line $\gamma = (\gamma_1, \dots, \gamma_n) : [0, 1] \to \mathbb{R}^n$ and $\gamma(\alpha) = X' + \alpha(X - X')$ where $X' \in \mathbb{R}^n$ is a black (all zero) image, $X \in \mathbb{R}^n$ is the test input and $\alpha \in [0, 1]$. Then the line integral is

$$IG_i(X) := \int_{\alpha=0}^1 \frac{\partial f^{(c)}(\gamma(\alpha))}{\partial \gamma_i(a)}\frac{\partial \gamma_i(\alpha)}{\partial \alpha} d\alpha\,$$

with $f^{(c)} : \mathbb{R}^n \to [0, 1]$ being the neural network. In practice, the integral is approximated by the Riemann sum. \emph{Blur Integrated Gradients} \cite{DBLP:journals/corr/abs-2004-03383} proposed to replace the black baseline by filtering the image with the Gaussian blur filter. \emph{Guided Integrated Gradients} \cite{DBLP:journals/corr/abs-2106-09788} enhanced the straight line path by an adaptive path that depends on the model.

\textbf{Class Activation Maps (CAM)} can be considered as variants of \emph{Gradients} in most cases. Instead of computing the gradient with respect to the input $X$, the back-propagation step is stopped at a chosen layer. For example, the {\it attribution map} for the popular {\it GradCAM}  \cite{Selvaraju_2019} is given by 
$$GradCam(X) := ReLU\left(\sum_k w_k^{(c)} A_k\right)\,,$$ with $w_k^{(c)} = \frac{1}{H \cdot W} \sum\limits_{i=1}^H \sum\limits_{j=1}^W        \frac{\partial f^{(c)}}{\partial A_k(i, j)}$ and where $H$ is the image height, $W$ the image width, and $A$ is the activation of the chosen layer. It is important to note, that due to the pyramid architecture of CNNs, the resolution of CAMs is typically much lower than the input images. Hence, CAMs need to be up-sampled, or inputs have to be down-sampled to compute an AM visualization.      
{\it GradCAM++} \cite{Selvaraju_2019} improved the original by weighting the gradients differently \cite{Chattopadhay_2018}. {\it Smooth GradCAM++} \cite{DBLP:journals/corr/abs-1908-01224} adds additional Gaussian noise to the input, like {\it SmoothGrad}. Finally, {\it LayerCAM} \cite{9462463} and {\it XGradCAM} \cite{DBLP:journals/corr/abs-2008-02312} also provide a different weighting schemes for the computed gradients.

Unlike the previous methods, {\it ScoreCAM} \cite{wang2020scorecam} does not require the computation of gradients (and is therefore not a variant of Gradients). The importance of channels in an intermediate layer is given by the confidence change when parts of the activation values are removed.

There are countless other {\it attribution map} methods that are improvements, combinations and extensions of the aforementioned methods \cite{DBLP:journals/corr/abs-1711-06104} such as {\it SS-CAM} \cite{https://doi.org/10.48550/arxiv.2006.14255}, {\it IS-CAM} \cite{DBLP:journals/corr/abs-2010-03023}, {\it Ablation-CAM} \cite{9093360}, {\it FD-CAM} \cite{https://doi.org/10.48550/arxiv.2206.08792}, {\it Group-CAM} \cite{DBLP:journals/corr/abs-2103-13859}, {\it Poly-CAM} \cite{https://doi.org/10.48550/arxiv.2204.13359}, {\it Zoom-CAM} \cite{DBLP:journals/corr/abs-2010-08644}, {\it Recipro-CAM} \cite{byun2023reciprocam} and {\it EigenCAM} \cite{DBLP:journals/corr/abs-2008-00299}.

\textbf{Black-box methods.} Black box methods rely on masking the input in some way, e.g. randomly occluding the input image and recording the change in the class probabilities. A notable example is {\it RISE} \cite{https://doi.org/10.48550/arxiv.1806.07421} and its precursor introduced in \cite{DBLP:journals/corr/ZeilerF13}. Other examples are \cite{DBLP:journals/corr/abs-1910-08485,DBLP:journals/corr/FongV17,DBLP:journals/corr/abs-1806-07421,DBLP:journals/corr/RibeiroSG16}.

\subsection{Evaluation of attribution methods}\label{sec:eval}
Given the large number of available attribution methods introduced in the previous section, users are left with the question which AM method to choose for their application (see \cref{fig:zero}). However, most publications that introduce novel AM methods are providing only qualitative visualizations of their results, showing selected input samples in direct comparison with other methods. Up to date, there is no commonly accepted benchmark that would provide a quantitative metric or ranking of {\it attribution map} algorithms for a given task. In the following, we review the few existing prior works.
Formally, we define a score function for the evaluation of AMs as $g(h, f, X) = o$ for $o \in \mathbb{R}$ with attribution method $h$, CNN $f$ and input $X$. For notational convenience, we leave out the superscript of the class $c$ and only consider a single image.

\textbf{\textit{Average Drop} (AD) and \textit{Increase in Confidence} (IIC)}. \textit{Average drop} has been suggested by \cite{Chattopadhay_2018}. It is defined as $g(h, f, X) := \max(f(X) - f(X \circ h(X)), 0)$, where $f(X)$ is the model’s \emph{softmax} output score (confidence) for class $c$. The operator $\circ$ denotes element-wise multiplication. The output $f(X \circ h(X))$ is the confidence of the model in class $c$ with only the so-called {\it explanation map} as input. The {\it explanation map} is derived by multiplying the input with the {\it attribution} map, such that unimportant pixels are hidden.

{\it Increase in confidence} is similarly defined as $g(h, f, X) := \mathbf{1}\left[f(X) < f(X \circ h(X))\right]$ where $\mathbf{1}[\cdot]$ is the indicator function.
Both methods quantify how much the output probabilities decrease or increase when given only the most salient regions in the input image.

\textbf{\textit{Complexity} (CP), \textit{Coherency} (CH) and \textit{ADCC}}. \cite{DBLP:journals/corr/abs-2104-10252} suggested measuring the number of pixels that are highlighted in a {\it attribution map}. They define $g(h, f, X) := ||h(X)||_1$. The basic idea follows the argument that a good {\it attribution map} should highlight only a small area.

{\it Coherency} is another evaluation score by the same authors. It measures how much the original {\it attribution map} differs from the \emph{attribution map} of the {\it explanation map}. It is defined by the correlation coefficient $g(h, f, X) := \frac{\text{Cov}(h(X), h(X \circ h(X)))}{\sigma_h\sigma_{hh}}$, where $h(X)$ is the input image {\it attribution map} and $h(X \circ h(X))$ is the {\it attribution map} of the {\it explanation map}.
Finally, the authors combine average drop, complexity and coherency to a single score by using the harmonic mean $\text{ADCC} := 3\left(\frac{1}{\text{CH}} + \frac{1}{1 - \text{CP}}+ \frac{1}{1 - \text{AD}}\right)^{-1}$. CP and CH have several drawbacks as evaluation functions. An attribution map with a few highlighted pixels is not always optimal. For example, if we look at an empty image, CP would always say that this map is optimal. Similarly, it is easy to construct counterexamples for CH where the score is high but the AM is meaningless. Since ADCC is only a combination of these metrics, this function will suffer from the same limitations.

\textbf{\textit{Deletion} and \textit{Insertion}}. Our approach for defining an evaluation function for saliency maps is based on these two metrics. They were introduced by \cite{DBLP:journals/corr/FongV17} and \cite{DBLP:journals/corr/abs-1806-07421}. These metrics will be expounded upon in detail in the next section. Del/Ins can be seen as generalizations of AD and IIC. AD and IIC evaluate the probability only once, while Del/Ins consider the input image as a sequence of images $I_1, I_2, \dots, I_n$ and calculate the change in the probability of the whole modification process.

\textbf{Other evaluation methods}. {\it Sparsity} \cite{DBLP:journals/corr/abs-2201-13291} follows a similar idea as {\it Complexity}. The authors also suggest using the correlation coefficient instead of AUC for the computation of the deletion and insertion scores. \\
Another approach is the {\it pointing game} \cite{DBLP:journals/corr/Zhang0BSS16}, with a variation introduced in \cite{https://doi.org/10.48550/arxiv.2208.06175}. The basic idea is to verify whether the pixel with the highest saliency score is contained in the bounding box of a specific object. 

\section{A Perturbation-Based Evaluation Method}

\begin{figure*}[ht]
  \centering
  \includegraphics[scale=0.5]{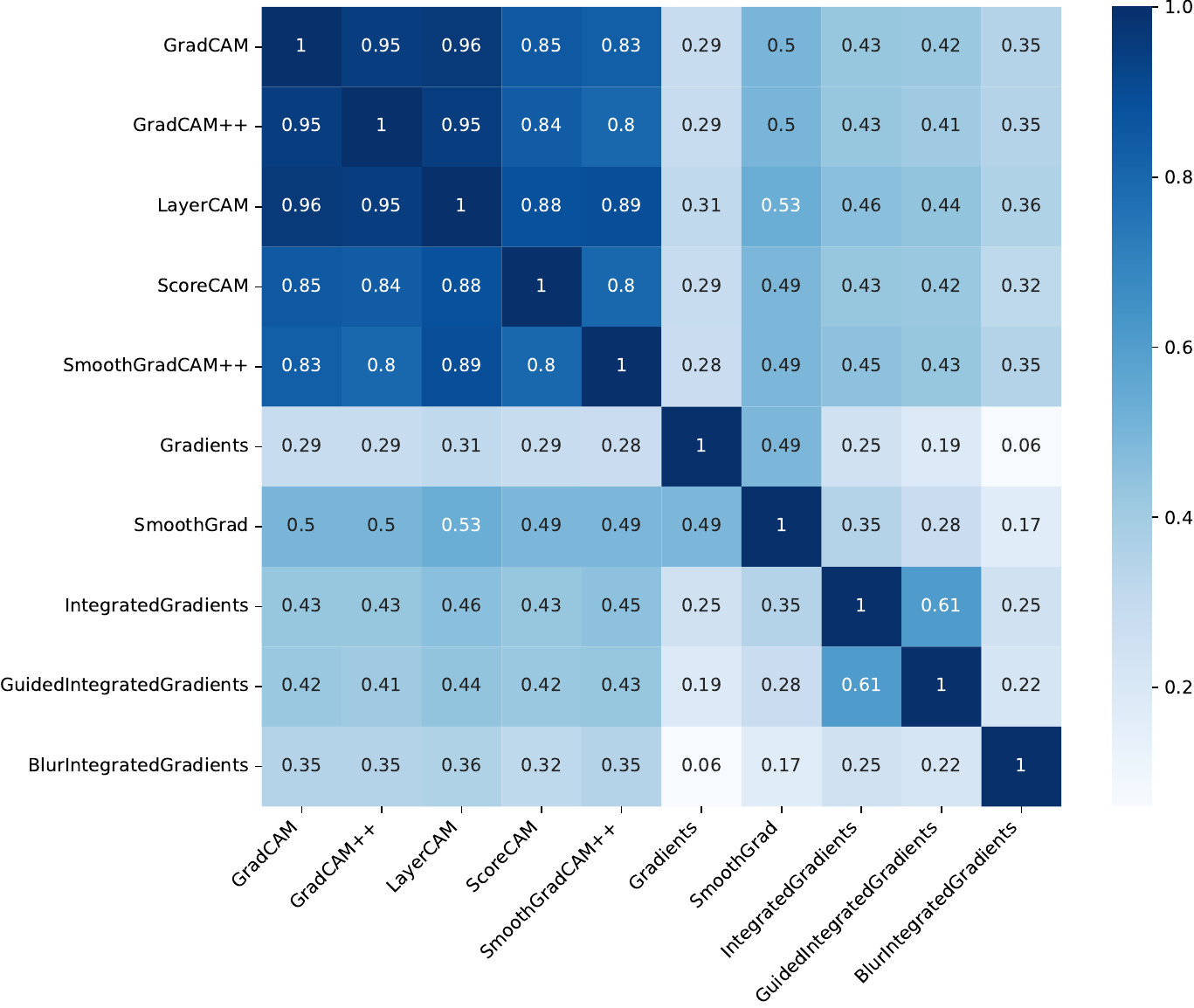}
  
   \caption{This plot illustrates the degree of similarity among all attribution maps. The matrix was computed by averaging the individual correlation results across all attribution maps in the ImageNet dataset with ResNet-50.}
   \label{fig:corr}
\end{figure*}

First, let us establish the need for introducing an evaluation function. As we have visually demonstrated in \cref{fig:zero}, saliency maps can exhibit significant variations. To precisely quantify these differences, we evaluate the similarity between individual \emph{attribution maps} (AM). To achieve this, we utilize the Pearson correlation coefficient, defined as:

$$r_{xy}={\frac {\sum _{i=1}^{n}(x_{i}-{\bar {x}})(y_{i}-{\bar {y}})}{{\sqrt {\sum _{i=1}^{n}(x_{i}-{\bar {x}})^{2}}}{\sqrt {\sum _{i=1}^{n}(y_{i}-{\bar {y}})^{2}}}}}\,,$$

where $x_i$ is the $i$th pixel in the AM and $\bar{x}$ is the sample mean. Similarly, $y_i$ is the $i$th pixel of the second AM. The output range of $r_{xy}$ is $[-1, 1]$, with $1$ indicating the highest similarity and $\leq 0$ a low similarity. $r_{xy} = 0$ is intuitively random noise and $r_{xy} = -1$ an "inverted" saliency map.

We assessed the similarity across a subset of 1000 images from ImageNet, employing both the ResNet-50 and ConvNeXt-tiny architectures. In \cref{fig:corr}, we present the results for ResNet-50, revealing an average similarity of 48\%. Remarkably, for ConvNeXt-tiny, we observed an even lower similarity of 22\%.

Given the substantial variation among these attribution maps, it becomes imperative to define a metric for evaluating which AMs most accurately represent what CNNs are truly "seeing".

Defining an evaluation function for {\it attribution maps} is inherently difficult because the actual ground truth is not known. Some publications \cite{9462463} on AM methods use segmentation masks from PASCAL VOC \cite{Everingham2009} or COCO \cite{DBLP:journals/corr/LinMBHPRDZ14} as pseudo ground truth to evaluate their results. However, we strongly argue against this practice since it imposes an AM score based on which regions CNNs {\it should} base its decisions on, whereas the task of AMs is to show on which regions it actually does rely on. For example, if a model dominantly uses the image background for its predictions, AM should reveal this flaw instead of trying to segment the foreground objects.

\begin{figure*}[ht]
  \centering
  \includegraphics[scale=0.14]{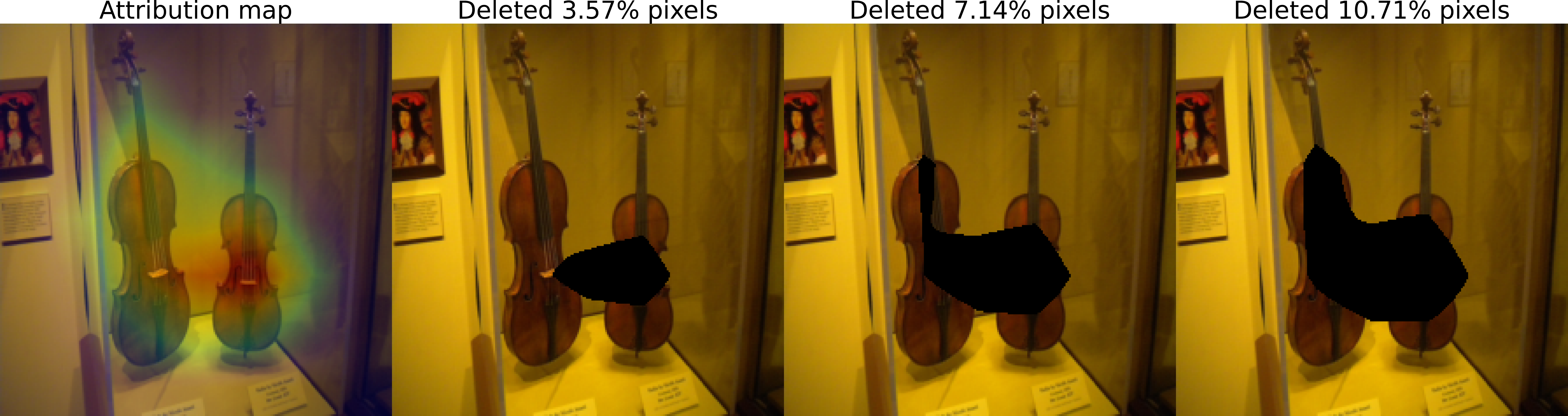}
  
   \caption{The series of images showcases the process of the Deletion methods, applied to the original image. The first image combines the original image with the saliency map overlay. Subsequent images (2nd to 4th) depict three stages of deletion, with progressively less important regions being zeroed out, as indicated by the colors in the first image.}
   \label{fig:deletionprocess}
\end{figure*}

More sophisticated approaches compensate the lack of ground truth information by use of surrogate functions. Treating the given neural network as a black box, the surrogate functions are used to manipulate the model input based on the AM under evaluation, deriving an AM score from the model response.

However, an underlying challenge emerges -- current surrogate functions induce a distribution shift in the image due to significant pixel alterations. This shift renders evaluation scores unreliable, often marking subpar saliency maps as superior, depending on the amount of alterations. Our proposed evaluation score overcomes these limitations inherent in previous metrics by rectifying the distribution shift. We first introduce the most prominent of these surrogate functions: the Deletion and Insertion score functions based on which we develop our evaluation measure.\\

\textbf{Deletion and Insertion.} The approach introduced by \cite{DBLP:journals/corr/FongV17} and \cite{DBLP:journals/corr/abs-1806-07421} revolves around the iterative deletion and insertion of pixels in input images. These pixels are chosen according to the highest values in the associated \emph{attribution map}. The deletion method commences with the original input image, and subsequently proceeds by iteratively masking sequentially significant regions, step by step, with a value of $0$. After each iteration step, we give the modified image to the neural network to generate a probability. \Cref{fig:deletionprocess} visually illustrates the deletion process. In contrast, for the insertion method, the procedure is inverted. Instead of directly introducing pixels into an initially empty image, a common variant of the insertion method first applies Gaussian blur to the image. This results in a smoothed version of the initial image which then serves as the starting point for the iterative insertion of pixels.

\begin{figure*}[ht]
  \centering
  \begin{subfigure}[b]{\textwidth}
    \centering
    \includegraphics[width=\textwidth, height=0.3\textheight, keepaspectratio]{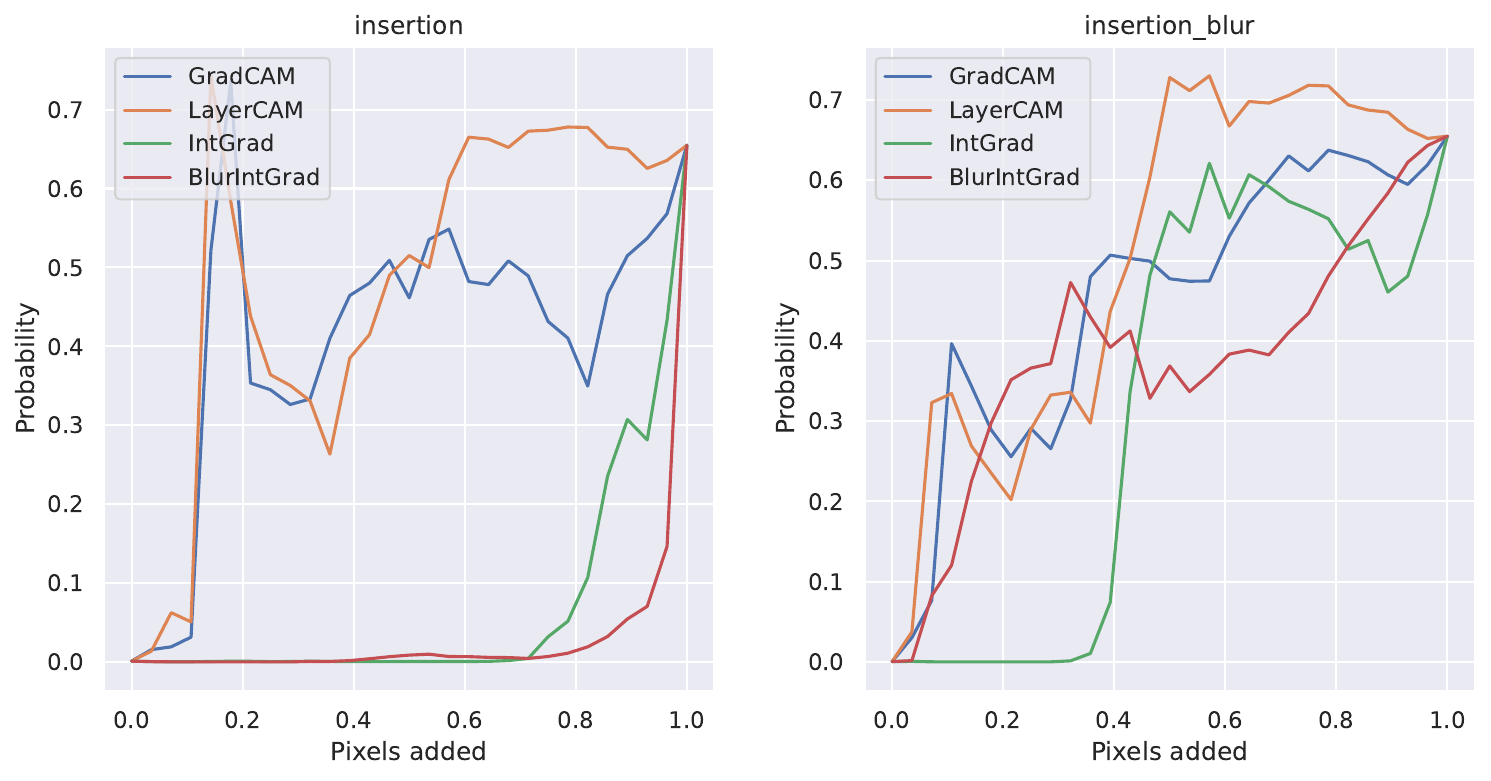}
    \caption{Insertion and Insertion Blur}
    \label{fig:insertionA}
  \end{subfigure}

  \begin{subfigure}[b]{\textwidth}
    \centering
    \includegraphics[width=\textwidth, height=0.5\textheight, keepaspectratio]{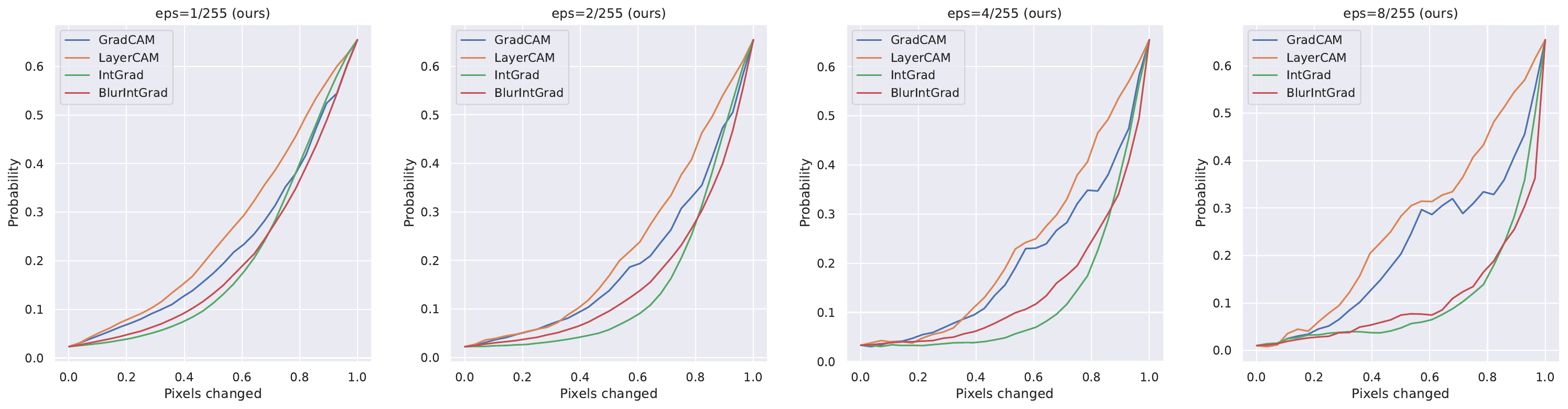}
    \caption{Our Perturbation method for Different Epsilon Values}
    \label{fig:perturbationA}
  \end{subfigure}
  
  \caption{Insertion, Insertion Blur, and our Perturbation method applied to one image. (a) Shows the Insertion and Insertion Blur methods. (b) Displays our Perturbation method for different epsilon values. Probability refers to the neural network's confidence for the selected class. The order of removing the perturbed pixels (b) and inserting the pixels (a) depends on the saliency map. The highest values are changed first. AUC values are computed for each saliency map, condensing the outcome into a single scalar value. Refer to \cref{tab:monosmooth} for numerical results using more than one image.}
  \label{fig:jointinsdelplot}
\end{figure*}

The outcome of this procedure can be graphically represented by plotting the number of inserted or deleted pixels along the x-axis and the neural network probability of the target class along the y-axis. \Cref{fig:insertionA} shows an example for the Insertion metric. When the image is black ("Pixels changed" is 0.0), the probability of the target class is zero. When all the original pixels are in the image, the probability is approx. 65\%.

The metric's evaluation is based on calculating the area under this plotted curve (AUC). For the "Deletion" metric, a good attribution map is indicated by a low AUC -- as we assume that all the most relevant pixels (represented by the map) for the models decision process have been masked out. The faster the probability for the ground truth class diminishes to zero, the more effective the map is considered to be. Conversely, in the case of the "Insertion" metric, the aim is for the probability to return fast to its original value, as only those pixels deemed most relevant by the saliency map are used for the models decision. Consequently, a high AUC indicates a good saliency map for the "Insertion" metric. In the evaluation section, we call the methods InsBlur (insertion with blurring), Ins (insertion without blurring) and Del (deletion).\\

\textbf{Adversarial perturbation.} The original motivation behind the Deletion/Insertion metric was to hide/emphasize the region responsible for determining the class. Each of the modified images is given to the CNN to produce a probability. When the probability of the ground truth class swiftly reaches zero/one, it signifies that the saliency map correctly identified the most crucial pixels for the original class decision. Nonetheless, an issue arises when setting pixels to zero: these types of images might not have been encountered during the neural network's training. Consequently, the CNN probability exhibits fluctuations (as seen in \cref{fig:insertionA}). The pixels could be misconstrued by the neural network as representing an additional object. To circumvent this, we propose concealing the significant regions within the image through minimal alterations. This approach enhances the score's reliability by maintaining a distribution closer to the original one.

To achieve this, we develop a gradient-based method using the FGSM adversarial attack. An adversarial attack's primary objective is to modify the image $X$ in the smallest possible way to a perturbed image $X^*$ that leads to a change in the models classification decision \cite{akhtar2018threat}. In other words, the class with the highest probability for the initial image $X$ should differ from the class with the highest probability for the perturbed image $X^*$.

Adversarial attacks can be categorized in various ways. Some attacks utilize gradient information to determine the necessary perturbations, while others rely solely on the model's class decision \cite{DBLP:journals/corr/RauberBB17}. Certain attacks do not even require knowledge of the true class label. Some approaches apply perturbations to all pixels, whereas others selectively target individual pixels for modification \cite{DBLP:journals/corr/abs-1710-08864}.

To define our metric for attribution maps, we use a gradient-based attack. This choice is driven by the substantial speed advantage of gradient-based attacks, as they maximize specific loss functions. It is noteworthy that our metric's effectiveness does not hinge on a perfect attack rate. Hence, stronger but slower attacks are not needed. We show in our ablation study that the results do not change when switching from the FGSM to the PGD attack.

Since we are concerned about distribution shifts, we apply perturbations uniformly to all pixels. This strategy introduces a milder distribution shift compared to selectively perturbing specific pixels by significant amounts. Reviewing the literature on detecting adversarial attacks, shows that certain stronger attacks such as the Carlini \& Wagner attack \cite{DBLP:journals/corr/CarliniW16a} are easier to detect by some methods \cite{8482346, DBLP:journals/corr/XuEQ17}. Additionally, one-pixel attacks such as \cite{DBLP:journals/corr/abs-1710-08864} cannot be used for defining a metric because for saliency maps potentially every pixel can be important. This suggests that small uniform perturbations are better suited for defining a metric.

Generating an adversarial example $X^*$ with a gradient-based attack can be described by the
following optimization problem \begin{equation} \arg\max_{\delta \in S} L(f(X + \delta), y)\,,\end{equation} where the perturbed samples are given by $X^* = X + \delta$, $L$ is the loss function, $y$ is the target class and $f(\cdot)$ is the neural network \cite{DBLP:journals/corr/abs-2111-09961}. The set of allowed perturbations is $S \in \mathbb{R}^d$, usually chosen to be $S = \{\delta \mid ||\delta||_{\infty} \leq \epsilon\}$ for a small $\epsilon \in \mathbb{R}$. This $\epsilon$ is not necessarily an infinitesimal change as in mathematics. The value is usually chosen to be $\frac{k}{255}$, where $k \in \{1, 2, \dots, 255\}$.

A solution to this gradient-based optimization problem using the $\ell_{\infty}$-norm for $S$ is the {\it fast gradient sign attack} (FGSM) \cite{https://doi.org/10.48550/arxiv.1412.6572}, where $\delta = \epsilon\cdot\text{sgn}\left(\frac{\partial L}{\partial X}\right)$ and $\text{sgn}\left(\cdot\right)$ is the sign function.

To each pixel of the original image either $-\epsilon$ or $\epsilon$ is added. When a pixel value is below $0$ or above $1$, the pixels are clipped back to the original domain. This does not affect the norm, since the strength of the attack is determined by the maximum. Therefore, the image attack's strength after clipping is still $\epsilon$.

It is important to note that adversarial attacks typically assume that images are normalized to the range $[0, 1]$, which simplifies the optimization problem by working in a continuous space. However, we want to specifically consider the case of discrete 8-bit images with values in the range $\{0, 1, \dots, 255\}$. In this context, the minimal perturbations that can be applied to an image are $1$ and $-1$.

We opt for discrete images to remain as close as possible to the original image distribution. When saving an image as a JPG file, all values become byte-valued, without fractional components. The FGSM attack mentioned earlier aligns with this requirement because it introduces $\pm 1$ perturbations for all pixels when $\epsilon = \frac{1}{255}$, satisfying the minimal perturbation criterion.

Other attacks do not necessarily satisfy the requirements of being discrete and uniform. For instance, the $\ell_{2}$-norm variant of FGSM with $\delta = \epsilon\cdot\frac{\partial L}{\partial X} \left(\|\frac{\partial L}{\partial X}\|_2\right)^{-1}$ is neither uniform nor discrete.\\



\textbf{Deletion/Insertion as $\ell_{\infty}$ adversarial attack}. Interestingly, we can view the original Deletion and Insertion metrics as specific instances of an $\ell_{\infty}$ adversarial attack. Consider the FGSM attack $\delta = \epsilon\cdot\text{sgn}\left(\frac{\partial L}{\partial X}\right)$ where $\frac{\partial L}{\partial X} < 0$ for all pixels of the image $X \in [0, 1]^{n\times m}$, and $\epsilon = \frac{255}{255}$. In this case, $\delta = -1$ and $X^* = \max\left(X + \delta, 0\right) = 0$. If the network interprets all input pixels as inversely correlated with the loss function, the entire image is blacked out. However, typically such high values $\epsilon$ are not employed, as adversarial attacks should not be perceptible to the human eye. Hence, if $\epsilon$ is a small value like $\frac{1}{255}$ or $\frac{8}{255}$, it would correspond to a slight increase or decrease in brightness.

These subtle brightness adjustments do not disrupt the output probability distribution (as shown in \cref{fig:perturbationA}). Moreover, color adjustments are a standard data augmentation technique during training. The only difference to the standard adversarial attack is that instead of changing the color globally, it is changed for each pixel individually. Hence, the attacks produce images, that are likely to have qualitatively occurred as training samples.\\

\textbf{Defining the score function.} We start with a fully perturbed image where to every pixel either $1$ or $-1$ was added with the FGSM attack. Gradually, this perturbation is systematically undone using the AM, as exemplified in \cref{fig:perturbationA}. As the perturbation is progressively reversed, we return to the original probability. However, when the network converges more rapidly to the initial probability, it signifies that the saliency map accurately pinpointed the pertinent regions. Hence, a high AUC value is essential for a saliency map to be deemed effective.

Unlike Deletion and Insertion, our functions are smooth and monotonic. This is obvious by comparing \cref{fig:insertionA} with \cref{fig:perturbationA}. Additional evidence is given in the evaluation section.

\begin{figure}[ht]
  \centering
  \includegraphics[scale=0.47]{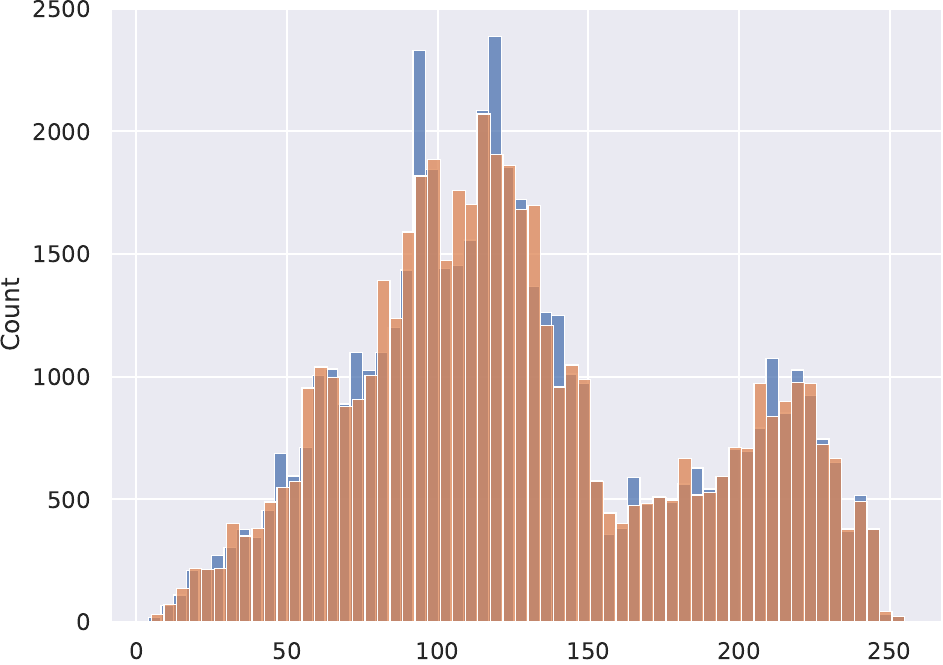}
  
   \caption{Blue is the original image, orange is the perturbed image. There is only a small effect on the image distribution. Increasing the brightness by $1$ would have a similar effect on the distribution (shifting it to the right). We considered here only the red channel of the violin image from \cref{fig:deletionprocess}.}
   \label{fig:effectonhistogram}
\end{figure}

Since we only change the pixels by $\pm 1$, it is the smallest conceivable attack with $\epsilon = \frac{1}{255}$. While it is possible to use higher $\epsilon$ values such as $\frac{8}{255}$ or $\frac{4}{255}$, this can also cause a distribution shift. We can see in \cref{fig:perturbationA} that higher epsilon values lead to less monotonic and smooth functions. In \cref{fig:effectonhistogram}, we see the effect of our perturbation on a histogram. The effect is as weak as increasing the brightness of an image by $1$.

The other metrics will always have a stronger distribution shift than our metric. If we look at the first iteration step of the Insertion metric, the entire image is initially black. A histogram would show a peak at 0 and nothing else, while our metric would still look like the original image.\\

\textbf{Other attacks.} We have motivated our adversarial perturbation with the FGSM attack. There are, however, $2^n$ other possible modifications for a given image (assuming $\pm 1$). Many of these perturbations (such as adding $-1$ to all pixels) do not affect the output probability of the neural network. FGSM also does not find for every image a probability-decreasing perturbation as it relies on a single gradient ascent step. An extension to FGSM uses {\it projected gradient descent} (PGD) \cite{https://doi.org/10.48550/arxiv.1706.06083} to strengthen the attack, but keeps the same $\delta$. Multiple iterations of FGSM are performed.

An important aspect of our evaluation is assessing the change in probability. Consequently, a precondition is the successful execution of the attack. Nonetheless, we have observed that weak attacks such as FGSM suffice for most datasets. It's not imperative to attain a 100\% attack success rate because we can skip images that were not successfully attacked.\\

\begin{table*}[ht]
  \centering
  \begin{tabular}{llllll}
    \toprule
        Attribution Method &
        Perturb (ours) $\uparrow$ & InsBlur $\uparrow$ & Del $\downarrow$ & Ins $\uparrow$ \\ \hline
GradCAM \cite{Selvaraju_2019} & 0.477 $\pm$ 0.08 & 0.786  $\pm$ 0.063 & 0.310 $\pm$ 0.148 & 0.694 $\pm$ 0.109\\
GradCAM++ \cite{Chattopadhay_2018} & 0.496 $\pm$ 0.106 & 0.772 $\pm$ 0.08 & 0.321 $\pm$ 0.168 & 0.679 $\pm$ 0.112\\
LayerCAM \cite{9462463} & 0.494 $\pm$ 0.094 & 0.766 $\pm$ 0.055 & 0.337 $\pm$ 0.162 & 0.666 $\pm$ 0.105\\
PolyCAMm \cite{https://doi.org/10.48550/arxiv.2204.13359} & 0.430 $\pm$ 0.098 & 0.720 $\pm$ 0.064 & 0.315 $\pm$ 0.206 & 0.532 $\pm$ 0.029\\
PolyCAMp \cite{https://doi.org/10.48550/arxiv.2204.13359} & 0.438 $\pm$ 0.110 & 0.720 $\pm$ 0.073 & 0.318 $\pm$ 0.208 & 0.542 $\pm$ 0.083\\
PolyCAMpm \cite{https://doi.org/10.48550/arxiv.2204.13359} & 0.440 $\pm$ 0.102 & 0.724 $\pm$ 0.064 & 0.320 $\pm$ 0.203 & 0.546 $\pm$ 0.077\\
ReciproCAM \cite{byun2023reciprocam} & 0.495 $\pm$ 0.137 & \textbf{0.794} $\pm$ 0.087 & 0.273 $\pm$ 0.149 & \textbf{0.708} $\pm$ 0.131\\
ScoreCAM \cite{wang2020scorecam} & 0.492 $\pm$ 0.113 & 0.764 $\pm$ 0.085 & 0.337 $\pm$ 0.177 & 0.661 $\pm$ 0.153\\
SmoothGradCAM++ \cite{DBLP:journals/corr/abs-1908-01224} & 0.441 $\pm$ 0.017 & 0.708 $\pm$ 0.040 & 0.427 $\pm$ 0.188 & 0.598 $\pm$ 0.115\\
\midrule
BlurintegratedGradients \cite{DBLP:journals/corr/abs-2004-03383} & 0.454 $\pm$ 0.136 & 0.758 $\pm$ 0.097 & \textbf{0.266} $\pm$ 0.353 & 0.521 $\pm$ 0.096\\
Gradients \cite{DBLP:journals/corr/SimonyanVZ13} & 0.544 $\pm$ 0.137 & 0.638 $\pm$ 0.152 & 0.293 $\pm$ 0.314 & 0.371 $\pm$ 0.237\\
GuidedintegratedGradients \cite{DBLP:journals/corr/abs-2106-09788} & 0.452 $\pm$ 0.138 & 0.697 $\pm$ 0.116 & 0.294 $\pm$ 0.319 & 0.505 $\pm$ 0.121\\
IntegratedGradients \cite{DBLP:journals/corr/SundararajanTY17} & 0.488 $\pm$ 0.135 & 0.712 $\pm$ 0.128 & 0.295 $\pm$ 0.319 & 0.532 $\pm$ 0.124\\
SmoothGrad \cite{DBLP:journals/corr/SmilkovTKVW17} & \textbf{0.562} $\pm$ 0.148 & 0.707 $\pm$ 0.128 & 0.295 $\pm$ 0.319 & 0.561 $\pm$ 0.140\\
\midrule
Canny \cite{canny1986computational} & 0.315 $\pm$ 0.072 & 0.564 $\pm$ 0.199 & 0.332 $\pm$ 0.279 & 0.401 $\pm$ 0.211\\
Uniform & 0.281 $\pm$ 0.061 & 0.528 $\pm$ 0.240 & 0.302 $\pm$ 0.276 & 0.302 $\pm$ 0.276\\
        \bottomrule
        \end{tabular}
  \caption{Evaluation scores averaged across all datasets and architectures, where Del = deletion, Ins = insertion on a black image, InsBlur = insertion on a blurred image, Perturb = FGSM perturbation. The first five AM methods require upsampling, while the next five methods output an AM in input resolution. The table shows that Del and Ins do not pass the baseline test, as Canny and Uniform should have the worst values. However, this table is only an overview and sanity check. We show in our evaluation section that only the ranking of Perturb is reliable.}
  \label{tab:averagedscore}
\end{table*}

\textbf{Baseline AMs and ground truth.}
For evaluation purposes, we introduce a set of synthetic baseline AMs which allows us to perform some model independent sanity checks on AM evaluation methods. The first of these baselines is called \emph{uniform}, which emulates an equal importance of all AM pixels by sampling from the uniform distribution. Logically, evaluation methods should penalize such an input. The second baseline, which we call \emph{canny}, applies the Canny edge detector \cite{canny1986computational} to the image and treats it as AM. This comparison with an edge detector was proposed in \cite{NEURIPS2018_294a8ed2}.
All these baseline {\it attribution maps} should achieve worse evaluation scores than actual AMs.

As mentioned earlier, segmentation masks are occasionally employed as a form of "ground truth" for assessing AMs. Nonetheless, a notable concern arises from the fact that not all pixels within an object hold equal significance. This phenomenon becomes evident during the prediction of a segmentation model. The central point of an object holds the highest probability, with probabilities gradually diminishing towards the object's edges. Thus, presuming a uniform probability of 100\% across all segmented pixels would inaccurately reflect the decision of a neural network. This situation is even more complicated for classification models because even a single part of an object could be sufficient for determining the class.\\

\textbf{Other scoring functions.} As detailed in the section covering related work, several other scoring functions such as \emph{Average Drop}, \emph{Increase in confidence}, and \emph{Coherency} rely on explanation maps. These maps involve the multiplication of the input by the saliency map. When a large portion of an image is blacked out, this induces the same distribution shift as the Deletion function. Consequently, they share the same unreliability as many other evaluation measures. In the evaluation section, we give evidence that any score function relying on masking the original image with zeros leads to unreliable results.

\section{Evaluation}

Given the absence of a definitive ground truth for saliency maps, the quantitative evaluation of our score function is based on consistency, mathematical properties, and comprehensive ablation studies.

To establish this, we conducted evaluations across various datasets and architectures to assess the robustness of the results. Unlike common practices in the literature, which often focus on large-scale datasets like ImageNet, we chose to incorporate real-world datasets, bolstering the credibility of our findings. Our evaluation encompassed three distinct datasets: ImageNet \cite{5206848}, Oxford-IIIT Pet Dataset \cite{parkhi12a}, and ChestX-ray8 \cite{WangPLLBS17}.

For all datasets, we analyze five CNN architectures: ResNet-50 \cite{DBLP:journals/corr/HeZRS15}, EfficientNet-B0 \cite{DBLP:journals/corr/abs-1905-11946}, DenseNet-121 \cite{DBLP:journals/corr/HuangLW16a}, ConvNeXt-Tiny \cite{DBLP:journals/corr/abs-2201-03545}, and RepVGG-B0 \cite{DBLP:journals/corr/abs-2101-03697}. The pre-trained weights were used for ImageNet, while the models were trained for the other datasets. This extensive assessment covers a total of 15 unique dataset-model combinations, constituting the most comprehensive comparison of saliency maps across diverse datasets to date.

Our evaluation implementation is based on the open {\it TorchCAM} \cite{torcham2020} and {\it Saliency Library} \cite{saliencylib} projects.

\subsection{Quantitative evaluation}

Each dataset-model combination (e.g. DenseNet-121 with ImageNet) defines a ranking. There are in total 16 attribution maps and 15 dataset-model combinations. Since we cannot show all 15 tables, we only show the averaged results of the 15 tables in
\Cref{tab:averagedscore}. However, we can analyze how the ranking across the different tables behaves numerically.\\

\textbf{Baselines.} Uniform and Canny serve as baseline methods designed to yield the lowest possible scores.

Across all 15 tables, we counted how often "Canny" was in second-last place and "Uniform" in last place (refer to \cref{tab:baselinetest}).

\begin{table}[h]
  \centering
\begin{tabular}{lll}
        \toprule
        Method & Canny 2nd to last $\uparrow$ & Uniform last $\uparrow$\\\hline
Del & 3 & 2\\
Ins & 4 & 11\\
InsBlur & 12 & 12\\
Perturb (ours) & \textbf{15} & \textbf{15}\\
        \bottomrule
        \end{tabular}
  \caption{Number of cases in which Canny edge detector and uniform came last. The maximum achievable score is 15.}
  \label{tab:baselinetest}
\end{table}

Only our method passes the baseline test. InsBlur comes second but it fails for certain architectures.\\

\textbf{Consistency.} Given that attribution methods are typically conceived as general approaches, it is reasonable to anticipate a degree of similarity in their behavior across diverse datasets and architectures. Consequently, it is desirable for the metrics to yield a consistent ranking across various datasets. 

To assess the degree of similarity between rankings, the rank correlation \cite{Newson2002} is often used in the statistical literature. A widely used measure for this purpose is Kendall's $\tau$. For example, if one ranking is SmoothGrad $<$ Gradients $<$ GradCAM++ and another ranking is SmoothGrad $<$ GradCAM++ $<$ Gradients, we can use this correlation measure to quantify the similarity between these two rankings.

Here we use Kendall's $\tau$ to measure the similarity across all 15 tables, resulting in the creation of a $15 \times 15$ matrix that captures the pairwise correlations between the rankings. For example, a correlation entry in this matrix would be how similar "DenseNet121 + Imagenet" and "ResNet50 + Oxford Pets" are. We compute this $15 \times 15$ matrix for all four metrics: Del, Ins, InsBlur, Perturb.

Since it would be too large to display the entire matrix, we average the entries to get an idea of the similarity. The results can be seen in \cref{tab:consistency}. However, the complete results can be found in the appendix \ref{secA1}.

\begin{table}[h]
  \centering
\begin{tabular}{ll}
        \toprule
        Method & Kendall's $\tau$ $\uparrow$\\\hline
Del & 0.236 $\pm$ 0.409\\
Ins & 0.353 $\pm$ 0.431\\
InsBlur & 0.432 $\pm$ 0.314\\
Perturb (ours) & \textbf{0.466} $\pm$ 0.252\\
        \bottomrule
        \end{tabular}
  \caption{Consistency across all model-dataset combinations.}
  \label{tab:consistency}
\end{table}

It turns out that our metric is the most consistent of the four. Since we are comparing different architectures and datasets, it is logical that Kendall's $\tau$ is not exceptionally high.

To get a better idea of the consistency of our metric, we average the values across the architectures and look at the top 3 AMs in \cref{tab:top3}.

\begin{table}[htbp]
  \centering
  \begin{tabular}{@{}cccc@{}}
    \toprule
    Ranking & Oxford & ChestXray & ImageNet \\
    \midrule
    1 & SmoothGrad & SmoothGrad & SmoothGrad \\
    2 & Gradients & Gradients & Gradients \\
    3 & ReciproCAM & \textbf{IntGrad} & ReciproCAM \\
    \bottomrule
  \end{tabular}
  \caption{Top-3 saliency methods for the three datasets. Results are averaged across 15 tables. Only IntegratedGradients (IntGrad) differs in the ChestXray dataset.}
  \label{tab:top3}
\end{table}

We highlighted the only attribution map that is different. Upon closer examination of all 15 dataset-architecture combinations, we found that SmoothGrad consistently claims the top position in almost all our metric's rankings. However, we would like to point out that SmoothGrad (similar to Gradients) is sometimes subject to noise. We recommend ReciproCAM or GradCAM++ as a noise-free and parameter-free alternative when noise becomes a problem. The choice of the CAM still depends on the dataset and the architecture.\\

\textbf{Monotonicity and Smoothness.} A score function exhibiting monotonic increase implies that the changes made to the pixels have not led to a distribution shift. For instance, if an image is completely obscured with black pixels and then gradually unveiled, the probability should only increase and not decrease. While revealing unimportant pixels devoid of pertinent information is possible, they should not lead to an information loss. Hence, the probability should always stay either at the same level or be higher.

In \cref{fig:jointinsdelplot}, we have already shown for a single image that our method leads to much smoother curves. We now extend this analysis to multiple images by defining smoothness measures.

We measure for Del the percentage of monotonic decrease and for all other evaluation functions the percentage of monotonic increase as defined by

$$\text{Monotonicity} = \frac{\sum_i \mathbf{1}[f(x_{i+1}) \geq f(x_i)]}{n-1}\,,$$

where $1[\cdot]$ is the indicator function. Furthermore, we suggest to measure the smoothness of the functions. Various mathematical definitions of smoothness exist, most of which are not directly applicable to discrete data. In this context, we employ a definition tailored to our scenario. We compute the function's fluctuation by measuring the standard deviation of its first derivative (approximated by the forward difference). The definition is as follows:

$$\text{Smoothness} = \frac{1}{n-1}\sqrt{\sum_{i=1}^n \left(f\left(x_{i+1}\right) - f\left(x_{i}\right) - \overline{x}\right)^2}\,,$$

where $\overline{x}$ is the sample mean. The motivation for measuring smoothness is exactly the same as for monotonicity. Smoothness also implies that probability cannot fluctuate at random.  As illustrated in \cref{tab:monosmooth}, our metric outperforms others in both measures, highlighting its robustness.

\begin{table}[ht]
  \centering
\begin{tabular}{lll}
        \toprule
        Method & Monotonicity $\uparrow$ & Smoothness $\downarrow$\\\hline
Del & 0.648 & 1.712\\
Ins & 0.671 & 1.658\\
InsBlur & 0.654 & 1.238\\
Perturb (ours) & \textbf{0.967} & \textbf{0.891}\\
        \bottomrule
        \end{tabular}
  \caption{Monotonicity and Smoothness across datasets and architectures.}
  \label{tab:monosmooth}
\end{table}

\begin{figure}[H]
  \centering
  \includegraphics[scale=0.45]{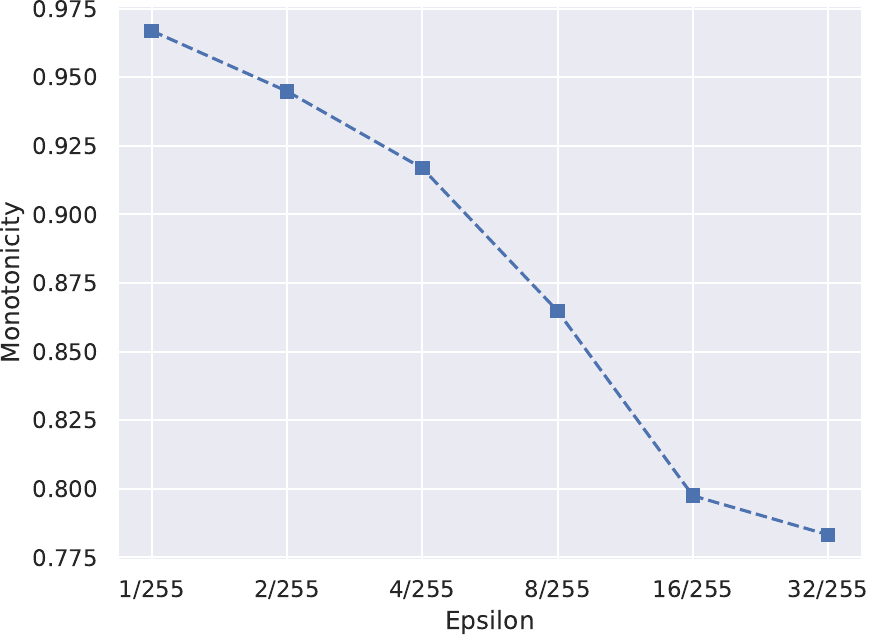}
  
   \caption{This plot shows that our score function is almost always increasing at $\epsilon = \frac{1}{255}$. But with stronger $\epsilon$, we would see fluctuations, similar to those seen with Insertion/Deletion.}
   \label{fig:monosmooth2}
\end{figure}

We also want to investigate what happens if we strengthen the adversarial attack by increasing $\epsilon$. For this purpose, we evaluate the monotonicity of our metric with respect to $\epsilon$ as seen in \cref{fig:monosmooth2}.

The results for smoothness are similar. We therefore place \cref{fig:smooth} in the appendix.\\

\begin{figure*}[ht]
  \centering
  \includegraphics[scale=0.55]{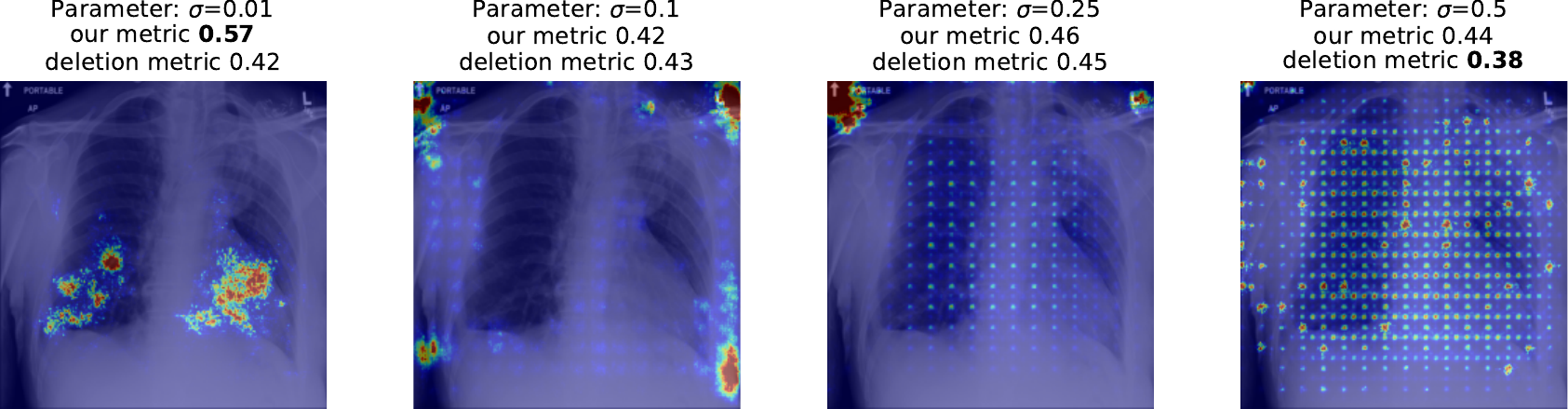}
  
   \caption{The optimal $\sigma$ values for SmoothGrad are determined by two evaluation methods: Deletion ($\downarrow$) and our metric ($\uparrow$). While visual analysis clearly suggests the first image as the most appropriate choice, the Deletion metric unexpectedly favors the last image as the preferred option.}
   \label{fig:application}
\end{figure*}

\textbf{Different adversarial attacks.} Adopting a stronger adversarial attack increases the number of successfully changed predictions for all images. However, if we consider the way the images are perturbed within the same gradient-based perturbation class, there is no difference. Consequently, switching to a stronger attack such as Projected Gradient Descent (PGD) does not lead to changes in the image distribution compared to Fast Gradient Sign Method (FGSM). We systematically repeated all experiments with PGD and observed a higher success rate of the attack. For example, in the case of ChestX-ray8, the success rate increased from 95.80\% to 100\%. Since the perturbation remains limited to $\pm 1$, we did not observe any change in the ranking of our scoring function. We advise resorting to PGD only when the attack success rate of FGSM proves insufficient. Other perturbation classes, such as $\ell_2$, are not recommended due to their allowance of fractional pixel values, as previously discussed.

Finally, opting for a black-box attack over a gradient-based approach like FGSM comes with the drawback of potentially producing images that are more distant from the data manifold (even if the constraint is still $\pm 1$). FGSM/PGD, being a first-order approximation, shifts the image towards a local maximum \cite{madry2019deep}. Since the method makes use of the gradient of the loss function to determine the direction, it cannot move that far away from the original image.

\subsection{Qualitative Evaluation}

In \cref{tab:consistency}, we observed that the evaluation function "Del" exhibited the least consistent outcomes, whereas our function displayed the most consistent results. This suggests that substantial variations in scores between different methods should be visually apparent, given the pronounced differences in their scores. For instance, if an evaluation function claims that a saliency map composed of random noise is the best, this conclusion is evidently erroneous. We have already established comparatively in \cref{tab:averagedscore} that this is indeed the case for the Deletion function (Uniform is better than GradCAM).

We now aim to corroborate this visual misalignment more explicitly. To achieve this, we turn our attention to a real-world application. The SmoothGrad saliency map includes a parameter for introducing Gaussian noise. The same image is processed through the neural network multiple times, each time with added noise. However, this parameter necessitates fine-tuning to attain optimal results. Traditionally, the tuning process relies on visual assessment of images while adjusting the parameter. This approach, although commonly used, lacks scientific rigor. In contrast, we propose employing our proposed metric to optimize this parameter. If an evaluation function is working as intended, then both visual and numerical assessments should yield congruent parameter values.

The outcomes are presented in \cref{fig:application}. We evaluated four noise parameters: $\sigma \in \{0.01, 0.1, 0.25, 0.5\}$. Visually, it's evident that the first image most accurately captures the neural network's relevant features. Conversely, the third and last images display grid artifacts, indicating poor performance. Our metric selects the first image as optimal, in line with expectations. However, the Deletion metric (Del) favors the last image. Consequently, our metric also produces the anticipated qualitative outcomes.

\section{Discussion and Outlook}

This paper addresses the pressing challenge of evaluating attribution methods such as class activation maps (CAMs) within the context of convolutional neural networks, particularly focusing on the reliability and consistency of current evaluation measures. The observed lack of correlation in the ranking of attribution maps (AMs) across different datasets underscores the complexity of this task and the limitations of existing evaluation techniques.

Our research underscores that the unreliability of many current evaluation measures arises from the significant distribution shifts introduced by common methods like masking. These shifts disrupt the stability and accuracy of evaluation outcomes, raising the need for a more robust evaluation framework. To tackle this issue, we introduce a novel approach that replaces masking with adversarial perturbations. This innovative modification maintains the original image distribution, leading to more consistent and accurate ranking of attribution methods.

Our proposed approach not only enhances the reliability of ranking but also aligns closely with expected results, providing a more trustworthy evaluation process. The demonstrated ability of our method to yield anticipated low scores for baseline methods like the Canny edge detector further supports its credibility.

Moreover, the inclusion of diverse neural network architectures in our evaluation adds to the generalizability of our findings. This observation underscores the robustness and versatility of our methodology, which has practical implications across various real-world applications.

The introduction of our perturbation-based evaluation method brings the concept of monotonicity and smoothness to the evaluation of attribution methods. The absence of significant fluctuations in the probability, as shown in our paper, further contributes to the reliability of our approach.

In terms of future outlook, there are several avenues for further research. Exploring the applicability of our method to even more complex and diverse datasets and architectures could provide deeper insights into its robustness and generalizability. For example, transformers and other sequence-based architectures could be considered for further work.

In conclusion, this paper contributes a significant step forward in addressing the challenges of evaluating attribution methods for deep learning models. By introducing a novel perturbation-based approach that addresses distribution shifts and maintains the original image distribution, we provide a more reliable and consistent framework for assessing attribution methods. This advancement holds the potential to enhance the credibility and applicability of attribution methods across diverse domains and applications in the realm of machine learning.

\textbf{Limitations.} A limitation of both our method and AMs in general is the case when non-existence of objects determines the CNN decision. Consider the case when the class is assigned when some object is not being in the image. Then there is no region to attack, and there are no salient pixels. Both our evaluation function and the AM would fail. 
Another limitation relates to the number of saliency maps used in our study. While our experiments included 16 different attribution maps, it is within the realm of possibility to extend our evaluation to more black box methods as well.

\bibliographystyle{unsrtnat}
\bibliography{references}

\begin{appendices}

\section{Consistency}\label{secA1}

We computed for all four metrics the pairwise Kendall's $\tau$. This results in four $15 \times 15$ matrices.

\begin{figure}[ht]
  \centering
  \includegraphics[scale=1.0]{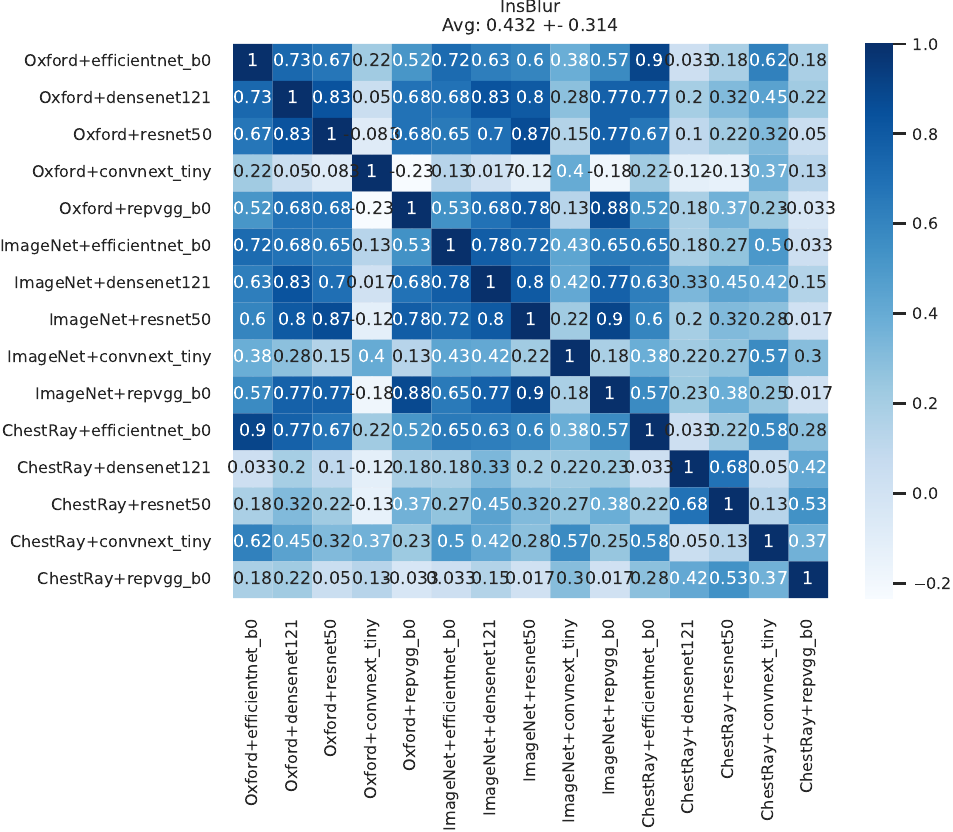}
  
   \caption{Correlation matrix for InsBlur.}
   \label{fig:cm1}
\end{figure}

\clearpage

\begin{figure}[ht]
  \centering
  \includegraphics[scale=1.0]{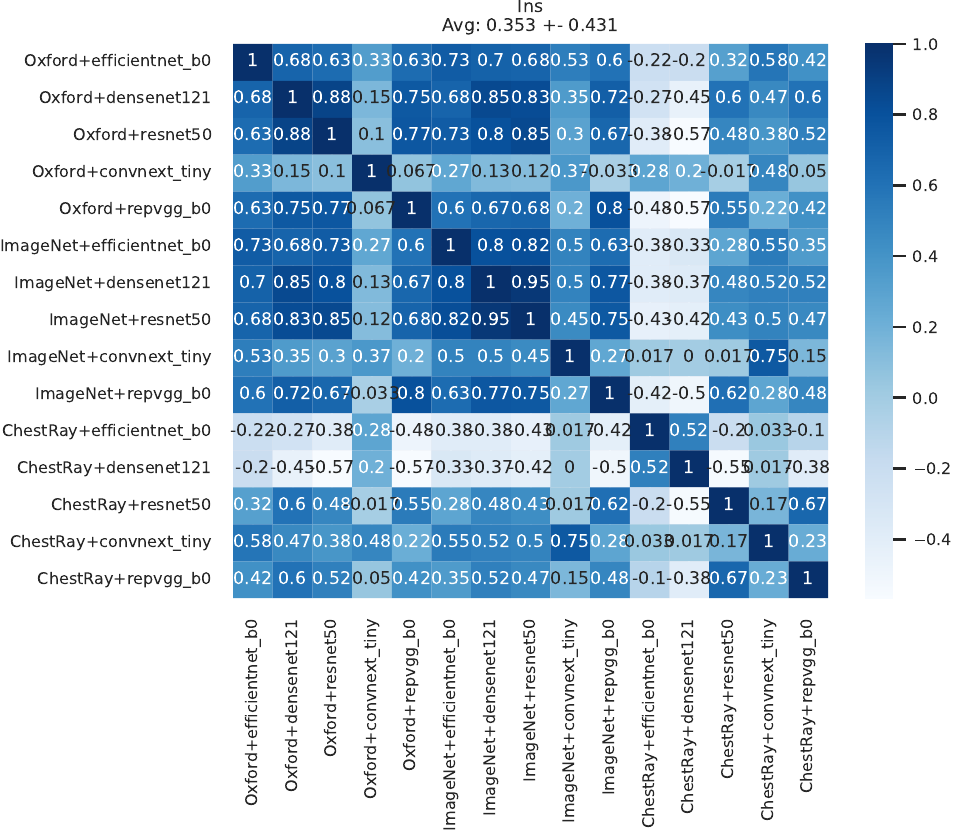}
  
   \caption{Correlation matrix for Ins.}
   \label{fig:cm2}
\end{figure}

\clearpage

\begin{figure}[ht]
  \centering
  \includegraphics[scale=1.0]{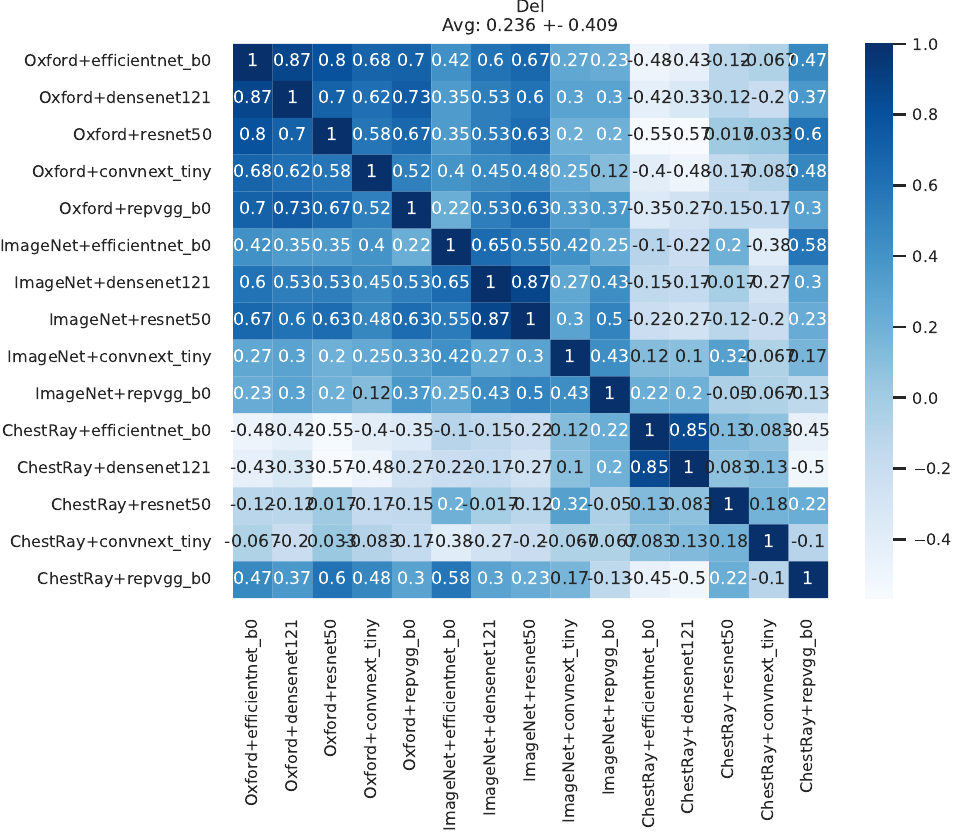}
  
   \caption{Correlation matrix for Del.}
   \label{fig:cm3}
\end{figure}

\clearpage

\begin{figure}[ht]
  \centering
  \includegraphics[scale=1.0]{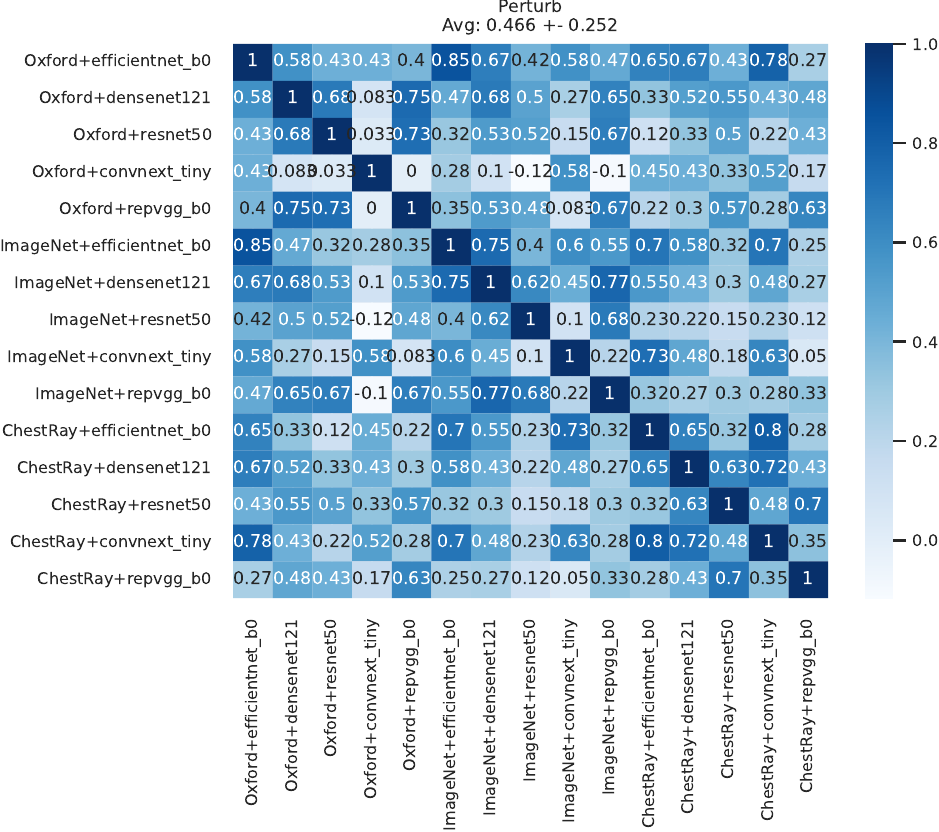}
  
   \caption{Correlation matrix for Perturb (ours).}
   \label{fig:cm4}
\end{figure}

\clearpage

\section{Smoothness}\label{secA2}

\begin{figure*}[ht]
  \centering
  \includegraphics[scale=1.0]{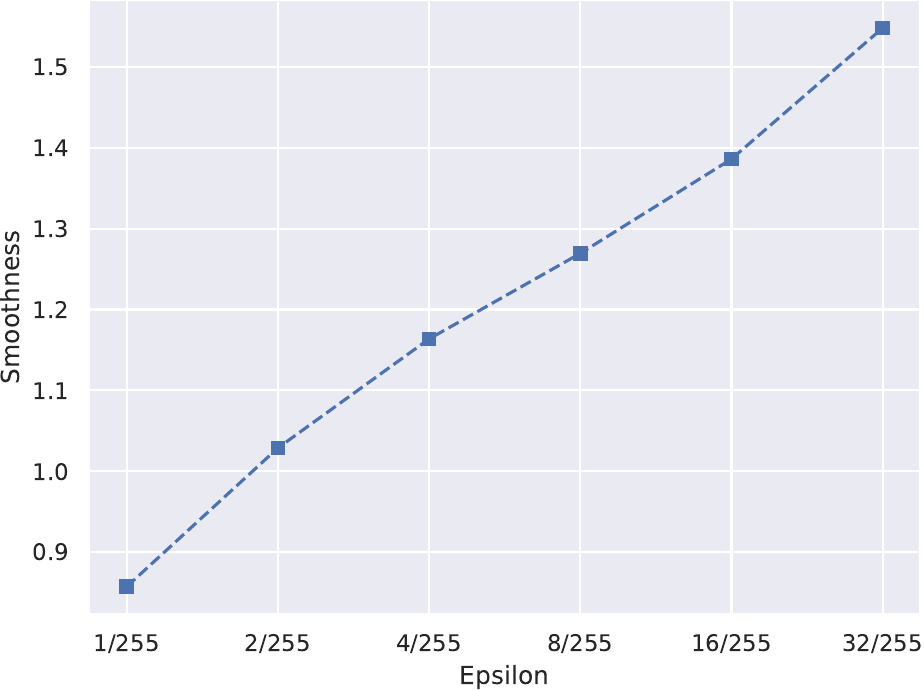}
  
   \caption{This plot shows the effect of $\epsilon$ on smoothness. Lower is better.}
   \label{fig:smooth}
\end{figure*}

The following \cref{fig:smooth} confirms the results that a lower $\epsilon$ is better.

\end{appendices}

\end{document}